\documentclass[sigconf,authordraft,screen]{acmart}
\usepackage{multirow}

\begin{document}

\title{
  % Prompt-agnostic Privacy Protection in Personalized Text-to-Image Synthesis
  Towards Prompt-robust Face Privacy Protection via Adversarial Decoupling Augmentation Framework 
  }

\author{
\textbf{Ruijia Wu}$^1$\,,
\textbf{Yuhang Wang}$^1$\,,
\textbf{Huafeng Shi}$^1$\,,
\textbf{Zhipeng Yu}$^2$\,,
\textbf{Yichao Wu}$^1$\,,
\textbf{Ding Liang}$^1$\,\\
$^1$\normalsize SenseTime Research \\
$^2$\normalsize School of Electronic, Electrical, and Communication Engineering, University of Chinese Academy of Sciences \\
}

\renewcommand{\shortauthors}{}
\renewcommand{\shorttitle}{}

\begin{abstract}
  Denoising diffusion models have shown remarkable potential in various generation tasks. The open-source large-scale text-to-image model, Stable Diffusion, becomes prevalent as it can generate realistic artistic or facial images with personalization through fine-tuning on a limited number of new samples. However, this has raised privacy concerns as adversaries can acquire facial images online and fine-tune text-to-image models for malicious editing, leading to baseless scandals, defamation, and disruption to victims' lives. Prior research efforts have focused on deriving adversarial loss from conventional training processes for facial privacy protection through adversarial perturbations. However, existing algorithms face two issues: 1) they neglect the image-text fusion module, which is the vital module of text-to-image diffusion models, and 2) their defensive performance is unstable against different attacker prompts. In this paper, we propose the Adversarial Decoupling Augmentation Framework (ADAF), addressing these issues by targeting the image-text fusion module to enhance the defensive performance of facial privacy protection algorithms. ADAF introduces multi-level text-related augmentations for defense stability against various attacker prompts. Concretely, considering the vision, text, and common unit space, we propose Vision-Adversarial Loss, Prompt-Robust Augmentation, and Attention-Decoupling Loss. Extensive experiments on CelebA-HQ and VGGFace2 demonstrate ADAF's promising performance, surpassing existing algorithms. 
\end{abstract}

\maketitle

\section{Introduction}
\label{sec:intro}
Recently, denoising diffusion models \cite{ho2020denoising, song2020denoising} have demonstrated remarkable potential in a wide range of generation tasks, including image synthesis, image inpainting, and multimodal tasks \cite{ilievski2017generative, wang2018unregularized}. Stable Diffusion \cite{rombach2022high}, the only currently available open-source text-based diffusion model, surpasses previous algorithms in terms of generation quality for text-to-image tasks.
Moreover, DreamBooth \cite{ruiz2022dreambooth} has been proposed to enable personalized generation by fine-tuning Stable Diffusion through few-shot learning. 
Owing to the outstanding generative capability, the applications of Stable Diffusion and DreamBooth have experienced rapid expansion.
Meanwhile, $diffusers$ \cite{von-platen-etal-2022-diffusers}, a widely popular diffusion model tool, emphasizes the importance of the image-text fusion module 
in text-to-image models. They also note that fine-tuning this module alone is sufficient for achieving excellent personalized generation results. As a result, they have introduced a fine-tuning functionality that exclusively updates the image-text fusion module, incorporating LORA \cite{hu2021lora}.

\begin{figure}[t]
  \centering
  \includegraphics[width=0.99\linewidth]{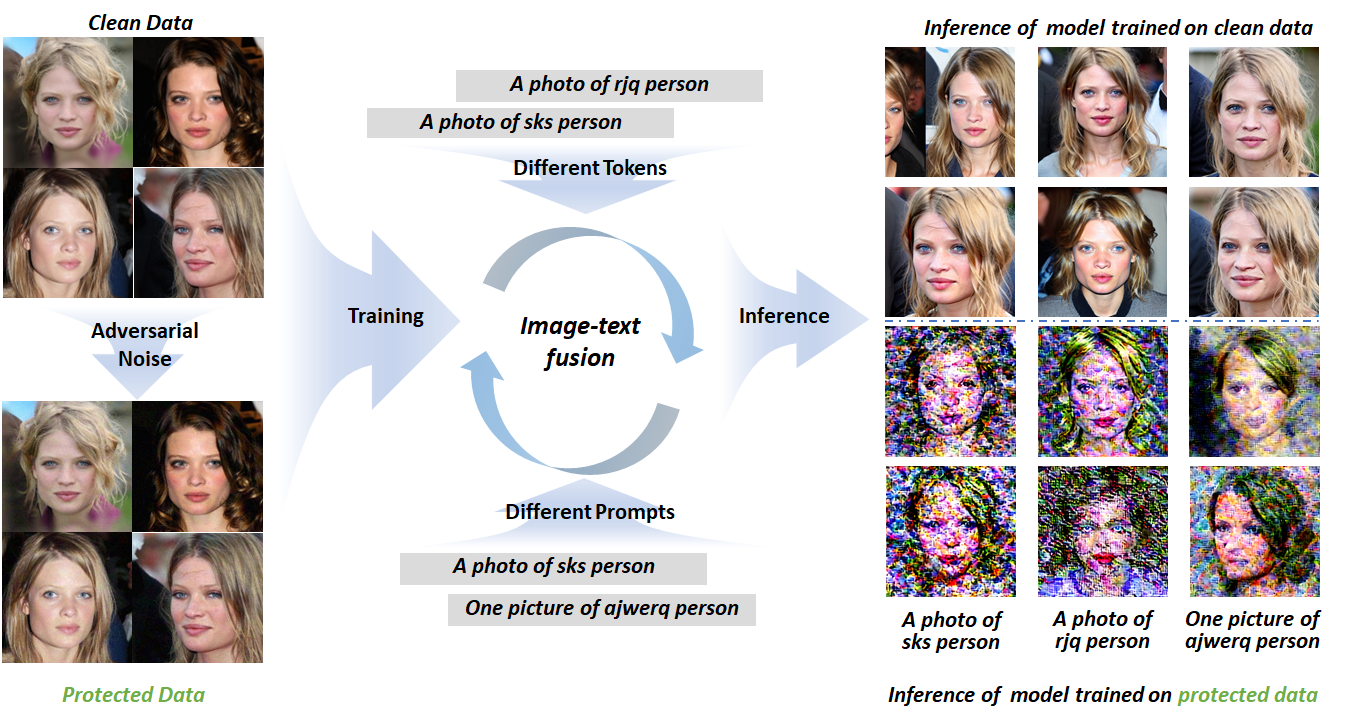}
  \vspace{-0.1in}
  \caption{Previous privacy protection methods neglect the vital module of text-to-image diffusion models, namely the image-text fusion module, and these methods are unstable against different attacker prompts. In this paper, we focus on the more challenging scenario, where 
 the adversary could utilize diverse prompts to fine-tune the text-to-image diffusion model.}
  \label{fig:fig1}
  \vspace{-0.15in}
\end{figure}

However, the exceptional generative capabilities of text-to-image models inevitably raise potential privacy concerns, such as facial privacy violations. Since DreamBooth can learn facial identity information by simply fine-tuning a few images, attackers can effortlessly acquire facial images online and fine-tune text-to-image models for malicious editing, resulting in baseless scandals, defamation, and disruption to victims' reputations and lives. 
Such attack ways can be categorized as DeepFakes \cite{juefei2022countering}, which represent one of the most severe AI crime threats that have garnered considerable attention from the media and community in recent years. Although the risks associated with GAN-based DeepFakes techniques are widely recognized and have stimulated significant research interest \cite{ge2021latent, yang2021defending, yeh2020disrupting}, 
the danger posed by diffusion models has yet to be fully acknowledged by the community.
Since current detection methods for facial images are insufficient in defending against harmful editing based on text-to-image diffusion models, it is essential to incorporate protective measures into images before release.

Currently, a few studies have also focused on this problem. 
Anti-DreamBooth\cite{van2023anti} devises an adversarial loss following the standard fine-tuning method (DreamBooth \cite{ruiz2022dreambooth}) for stable diffusion models, 
applying it to facial images to render protected sample IDs unlearnable for a personalized generation. 
However, the image-text fusion module, the essential component in text-to-image models, has been rarely studied. Additionally, \cite{van2023anti} utilizes a fixed text prompt during adversarial noise creation, impeding the maintenance of consistent defensive performance when the attacker employs varying prompts.

In this paper, we first focus on the vital component of text-to-image models, 
the image-text fusion module, 
by designing adversarial noise to enhance the defensive performance of facial privacy protection algorithms (shown in Fig. \ref{fig:fig1}). 
Furthermore, we introduce multi-level text-related augmentations to ensure the generated adversarial noise maintains defense stability when attackers employ various prompts for training or inference. Consequently, we propose the Adversarial Decoupling Augmentation Framework (ADAF), 
which disrupts image-text mapping through decoupling image-text fusion. 
Specifically, our ADAF designs effective strategies for vision, text, and the cross-modal common unit space.
For vision space, we propose Vision-Adversarial Loss (VAL), which approximates the output of text-to-image models to random noise, disrupting effective feature transformation in the image space. 
For text space, we introduce Prompt-Robust Augmentation (PRA), which reduces prompt information dependency from the text prompt level and enhances text feature diversity through random perturbations at the text embedding level. 
For the common unit space, we design Attention-Decoupling Loss (ADL) to disrupt the established image-text mapping relationships during the fusion process, effectively perturbing the core creative capabilities of text-to-image models. 
Extensive experiments on CelebA-HQ \cite{karras2017progressive} and VGGFace2 \cite{cao2018vggface2} demonstrate that our ADAF achieves promising performance with different attacker prompts and surpasses existing baselines. In general, our contributions are:
\begin{itemize}
    \item We introduce the Adversarial Decoupling Augmentation Framework, which is the first to generate adversarial noise targeting the decoupling of image-text fusion components to disrupt the training process of text-to-image models.
    \item At the vision space, we propose Vision-Adversarial Loss to perturb image features. At the text space, we introduce Prompt-Robust Augmentation to expand the perturbation text space at two levels. At the common unit space, we design Attention-Decoupling Loss to disrupt image-text mapping relationships.
    \item Extensive experiments demonstrate that our ADAF achieves promising performance with different attacker prompts and outperforms other methods.
\end{itemize}
\section{Related Work}
\label{sec:related}
\noindent\textbf{Text-to-image generation models}
The text-to-image generation model is a neural network that takes as input a natural language description and produces an image matching that description. With the emergence of new large-scale training datasets like LAION5B \cite{schuhmann2022laion}, 
there has been a rapid advancement in text-to-image generative models, 
which has garnered public attention and opened up new doors for visual-based applications.
In the realm of text-to-image generative models, there exists a diverse assortment of methodologies, encompassing auto-regressive \cite{yu2022scaling}, mask-prediction \cite{chang2023muse}, GAN-inspired \cite{sauer2023stylegan}, and diffusion-driven techniques. Each has demonstrated impressive qualitative and quantitative outcomes. Among these, diffusion-centric models excel at producing superior-quality, easily modifiable visuals, leading to their widespread adoption in text-to-image synthesis. GLIDE \cite{nichol2021glide} emerged as an early diffusion model employing classifier guidance for text-to-image generation. DALL-E 2 \cite{ramesh2022hierarchical} built upon this foundation by incorporating the CLIP text encoder and a diffusion-based prior, enhancing output quality. To optimize the balance between efficiency and fidelity, subsequent research either adopted a coarse-to-fine generation procedure, as in Imagen \cite{saharia2022photorealistic} and eDiff-I \cite{balaji2022ediffi}, or explored the latent space, exemplified by LDM \cite{rombach2022high}. StableDiffusion \cite{rombach2022high}, primarily influenced by LDM, became the first open-source large-scale model of its kind, bolstering the proliferation of text-to-image synthesis applications. These developments have propelled text-to-image generative models to achieve remarkable outcomes, unlocking new potential for visually-oriented applications.

\noindent\textbf{Adversarial attacks}
Adversarial attacks are inputs intentionally designed to
mislead deep learning models during inference but are imperceptible to human visions. In the seminal work on the Fast Gradient Sign Method (FGSM) attack \cite{goodfellow2014explaining}, adversarial vulnerability research in machine learning gained significant traction. 
The primary objective of such attacks is to craft model inputs that provoke misclassifications while retaining visual similarity to clean inputs. 
Post this groundbreaking contribution, numerous attacks employing diverse strategies surfaced, 
with more prominent ones, including: 
FGSM's iterative variants \cite{kurakin2018adversarial,madry2017towards}, 
implicit adversarial perturbation magnitude restriction via regularization 
as opposed to projection in \cite{carlini2017towards}. 
Moreover, the approach of locating a proximate decision boundary to breach in \cite{moosavi2016deepfool}, among others.
For black-box attacks, wherein the attacker lacks complete access to model weights and gradients, 
\cite{uesato2018adversarial} approximates gradients employing sampling techniques, 
while \cite{brendel2017decision,chen2020hopskipjumpattack,andriushchenko2020square} endeavor to 
concoct a nearby example by probing the classification 
boundary and subsequently identifying the direction for achieving a viable adversarial instance. 
As a composite of various attacks, \cite{croce2020reliable} serves as a commonly employed benchmark metric, 
capable of overcoming gradient obfuscation \cite{athalye2018obfuscated} through 
the expectation-over-transformation method \cite{athalye2018synthesizing}.
In this paper, we use the adversarial attack to generate adversarial noise, added to face images to implement facial privacy protection upon text-to-image models.

\noindent\textbf{Privacy protection with image cloaking}
With the rapid advancement of deep learning, several potential risks have been raised. Misusing AI models to exploit public images for malicious purposes gradually becomes a critical risk and has motivated researchers to propose the approach of "image cloaking", 
which involves adding subtle noise to users' images before publication to disrupt any attempts to exploit them.
Our proposed methods fall under this category.

Image cloaking is vital in protecting privacy against unauthorized facial recognition and counteracting GAN-based DeepFakes. Techniques like Fawkes \cite{shan2020fawkes} employ targeted attacks to alter identity within the embedding space, while Lowkey \cite{cherepanova2021lowkey} enhances transferability using surrogate model ensembles and Gaussian-smoothed perturbed images. AMTGAN \cite{hu2022protecting} generates inconspicuous cloaks through makeup transfer, and OPOM \cite{zhong2022opom} optimizes individualized universal privacy masks.
Approaches such as Yang et al. \cite{yang2021defending} leverage differentiable image transformations for robust cloaking, and Yeh et al. \cite{yeh2020disrupting} introduce novel objective functions to neutralize or distort manipulation. Huang et al. \cite{huang2021initiative} address personalized DeepFakes by alternating surrogate model training and perturbation generators. Anti-Forgery \cite{wang2022anti} designs perturbations for Lab color space channels to achieve natural-looking, robust cloaking. Recently, UnGANable \cite{li2022unganable} disrupts StyleGAN-based manipulation by obstructing inversion, contributing to image cloaking advancements.
Our method complements these works by disrupting image manipulation across a broad spectrum of generative models.
\section{Methodology}
\subsection{Preliminaries}
\label{sec:preliminaries}
% \noindent \textbf{Diffusion Models Based Text-to-image Generation.}
\noindent \textbf{Unconditional Diffusion Models.}
Diffusion models, a category of generative models \cite{ho2020denoising,sohl2015deep}, are defined by forward and backward processes. The forward process incrementally introduces noise to an input image until it reaches Gaussian noise. Conversely, the backward process reconstructs the desired data from random noise, reversing the forward process.
Considering an input image $x_0 \sim q(x)$, the diffusion process utilizes a noise scheduler ${\beta_t: \beta_t \in {0,1}}_{t=1}^T$ to apply increasing amounts of noise over T steps, resulting in a sequence of noisy variables: ${x_1, x_2, \dots, x_T}$. Each variable $x_t$ is generated by introducing noise at the respective timestep t:

\begin{equation}
  \begin{aligned}
    x_t=\sqrt{\bar{\alpha}_t} x_0+\sqrt{1-\bar{\alpha}_t} \epsilon
  \end{aligned}
  \label{eq:forward_process}
\end{equation}
where $\alpha_t=1-\beta_t$, $\bar{\alpha}_t=\prod_{s=1}^t \alpha_s$, and $\epsilon \sim \mathcal{N}(0, \mathbf{I})$. 
The backward process estimates the injected noise $\epsilon$ using a neural network $f_{\theta} (x_{t+1},t)$, and minimizes the $\ell_2$ distance between estimated and true noise:
% \vspace{-0.05in}
\begin{equation}
  \begin{aligned}
    \mathcal{L}_{u n c}\left(\theta, x_0\right)=\mathbb{E}_{x_0, t, \epsilon \in \mathcal{N}(0,1)}\left\|\epsilon-f_\theta\left(x_{t+1}, t\right)\right\|_2^2,
  \end{aligned}
  \label{eq:L_unc}
\end{equation}
where $t \in {1,...,T}$.

\noindent \textbf{Text-based Diffusion Models.}
The above description explains the working principle of an unconditional diffusion model. When conditions are required, the denoising process will change. This paper focuses on text-to-image diffusion models, which further control the sampling process with an additional prompt $c$ to generate photo-realistic outputs well-aligned with the text description. The objective is formulated as follows:
\begin{equation}
  \begin{aligned}
    \mathcal{L}_{\text {cond }}\left(\theta, x_0\right)=\mathbb{E}_{x_0, t, c, \epsilon \in \mathcal{N}(0,1)}\left\|\epsilon-f_\theta\left(x_{t+1}, t, c\right)\right\|_2^2.
  \end{aligned}
  \label{eq:cond}
\end{equation}
% \vspace{-0.05in}
With a prompt as a condition, text-to-image models can modify images based on the input prompt or generate content related to the prompt from scratch. Currently, the only open-source text-based large model is the stable diffusion, which we will use as the basis for our research.

\noindent \textbf{DreamBooth.}
DreamBooth is a fine-tuning approach aimed at customizing text-to-image diffusion models for specific instances of interest. This method pursues two primary goals. First, it compels the model to reconstruct the user's images using a generic prompt $c$, such as "a photo of sks [class noun]", where "sks" signifies the target user, and "[class noun]" denotes the object category, like "dog" for animal subjects. For training, DreamBooth employs the base loss of diffusion models as described in Eq. \ref{eq:db}, with $x_0$ representing each user's reference image. Second, it introduces a prior preservation loss to mitigate overfitting and text-shifting issues when a limited set of instance examples are available. Specifically, it utilizes a generic prior prompt $c_reg$, for example, "a photo of [class noun]", and enforces the model to reproduce instance examples generated randomly from that prior prompt utilizing the initial weights $\theta_ini$. The training objective is formulated as follows:
% \vspace{-0.1in}
\begin{equation}
  \begin{aligned}
    \begin{aligned}  \mathcal{L}_{DB}\left(\theta, x_0\right)&=\mathbb{E}_{x_0, t, t^{\prime}}\left\|\epsilon-f_\theta\left(x_{t+1}, t, c\right)\right\|_2^2 \\ 
      &+\lambda\left\|\epsilon^{\prime}-f_{\theta_{\text {ini }}}\left(x_{t^{\prime}+1}, t^{\prime}, c_{reg}\right)\right\|_2^2\end{aligned},
  \end{aligned}
  \label{eq:db}
\end{equation}
where $\epsilon$ and $\epsilon^{\prime}$ are both in $N(0,I)$ distribution and $\lambda$ is the weight of loss about prior preservation items.

\noindent \textbf{Privacy Protection by Adversarial Perturbation. }
The objective of adversarial attacks is to search for a barely perceptible perturbation of an input image that can mislead the behavior of specific models. Conventional research has focused on classification problems where, for a given model $f$, an adversarial example $x^{\prime}$ of an input image $x$ is generated to remain visually undetectable while causing a misclassification or targeted attack. The minimal visual difference is generally enforced by constraining the perturbation within an $\eta$ ball concerning an $\ell_p$ metric, i.e., $\left|x^{\prime}-x\right|_p<\eta$.

To safeguard against attackers who exploit generative models to create maliciously edited personal images, recent approaches like \cite{van2023anti} suggest incorporating similar adversarial perturbations to clean images to impede the training of generative models. Let $\mathcal{X}$ represent the set of images of the person to be protected. For each image $x \in \mathcal{X}$, \cite{van2023anti} introduces an adversarial perturbation $\delta$ and publishes the altered image $x^{\prime}=x+\delta$ while keeping the original image private. The published image set is $\mathcal{X}^{\prime}$. An adversary can gather a small set of that person's images, $\mathcal{X}_{{DB}^{\prime}}=\{{x^{(i)}+\delta^{(i)}\}}_{i=1}^{N_{DB}} \subset \mathcal{X}^{\prime}$. The adversary then utilizes this set as a reference to fine-tune a text-to-image generator $f_\theta$ by following the DreamBooth algorithm to obtain the optimal hyperparameters $\theta^*$. As the DreamBooth model overfits the adversarial images, it is deceived into performing poorly in reconstructing clean images. The overall training objective can be summarized as follows:

\begin{equation}
  \begin{aligned}
    \delta^{*(i)}&=\underset{\delta^{(i)}}{\arg \max } \mathcal{L}_{\text {ADV }}\left(f_{\theta^*}, x^{(i)}\right), \forall i \in\left\{1, \ldots, N_{DB}\right\}, \\
    \text { s.t. }& \quad \theta^*=\underset{\theta}{\arg \min } \sum_{i=1}^{N_{DB}} \mathcal{L}_{DB}\left(\theta, x^{(i)}+\delta^{(i)}\right), \\
    \text { and } &\quad\left\|\delta^{(i)}\right\|_p \leq \eta \quad \forall i \in\left\{1, \ldots, N_{DB}\right\},
  \end{aligned}
  \label{eq:antidb}
\end{equation}
where $\mathcal{L}_{DB}$ is defined in Eq. \ref{eq:db}, and $\mathcal{L}_{\text {ADV}}$ presents a challenge for us to design an appropriate adversarial loss, which can produce a robust noise, ensuring the face ID information remains unlearnable by attackers, regardless of the variety of prompts employed, thereby achieving the protection objective. In \cite{van2023anti}, $\mathcal{L}_{\text {ADV}}$ is composed of $\mathcal{L}_{\text {cond }}$ and an alternating training with function $f$. Nevertheless, it does not take into account the vital module in the text-to-image model, the image-text fusion module. Moreover, using fixed text prompts for poisoning renders the generated adversarial noise unable to maintain defense stability when attackers use different prompts. We will provide a detailed introduction to our $\mathcal{L}_{\text{ADV}}$ in Section \ref{sec:method}.

\begin{figure*}[]
  \centering

  \includegraphics[width=0.9\linewidth]{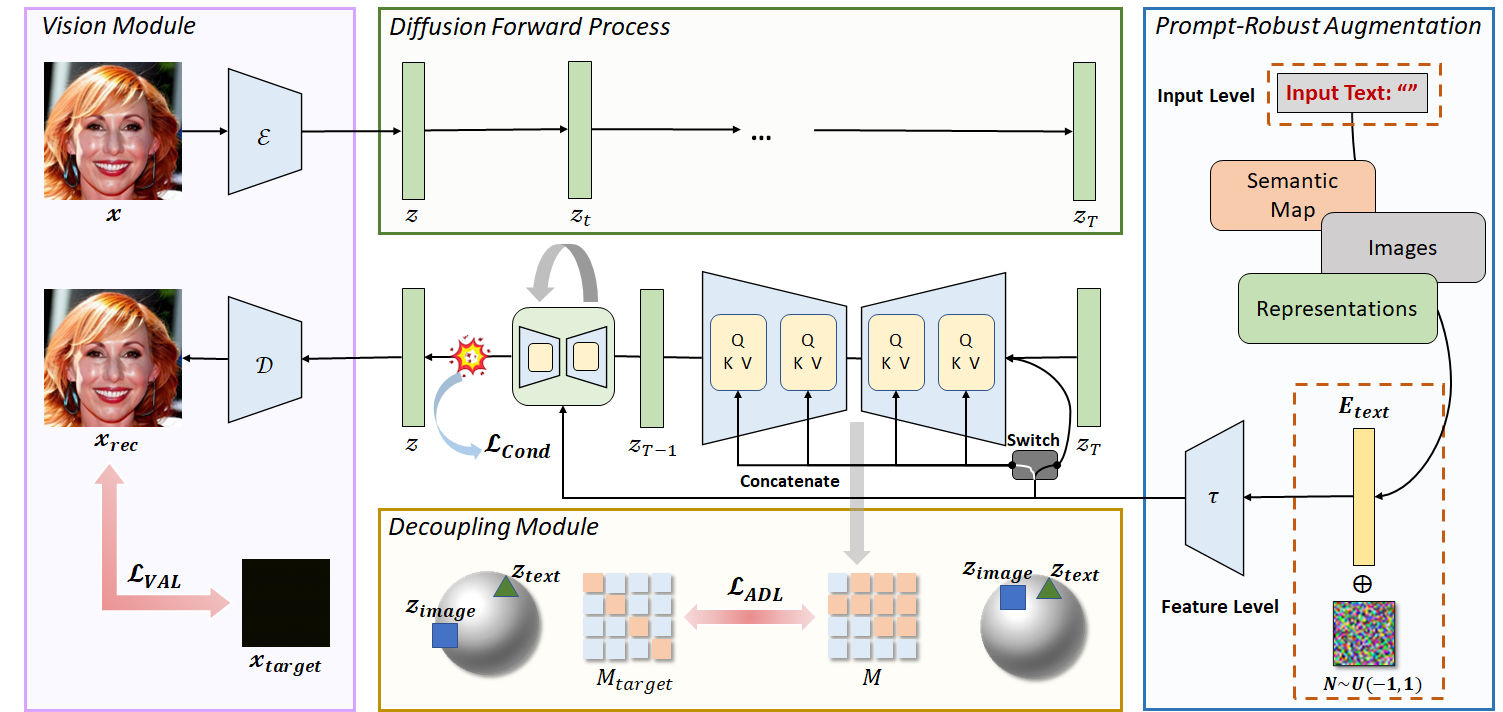}
  \caption{Overview of our ADAF. In the vision module, we propose Vision-Adversarial Loss ($\mathcal{L}_{VAL}$) to interfere with the vision space by bringing the model’s reconstructed image ($x_{rec}$) closer to the target image ($x_{target}$). In the decoupling module, we design Attention Decoupling Loss ($\mathcal{L}_{ADL}$), which aims to draw the attention matrix ($M$) close to a target matrix ($M_{target}$), thereby disrupting the attention matrix’s ability to express correct image-text correlation coefficients. Prompt-Robust Augmentation (PRA) is introduced to enhance defense stability against diverse attack prompts, containing two levels, the input and feature levels, respectively. In addition, we also considered the adversarial loss corresponding to the conventional DreamBooth training ($\mathcal{L}_{Cond}$).}
  \label{fig:fig2}
  % \vspace{-0.1in} 
\end{figure*}

\subsection{The Proposed Method}
\label{sec:method}
In this section, we present a detailed description of our Adversarial Decoupling Augmentation Framework (ADAF), which focuses on disrupting the critical image-text fusion module in text-to-image diffusion models and enhances the defense stability of protection algorithms against different attack prompts. Our ADAF is illustrated in Fig. \ref{fig:fig2}.

\noindent \textbf{Vision-Adversarial Loss (VAL)}
For image generation models, the ability to capture image features and reconstruct training images during the training process is fundamental and crucial. Effectively transforming extracted features from feature space to image space to generate images resembling the training samples indicates the model's efficient capability in the vision space. Based on this fact, we propose a disruption loss concerning the vision space as follows:
% \vspace{-0.1in}
\begin{equation}
    L_{VAL} = \left\| {x_{rec} - x_{target}} \right\|^{2},
    \label{eq:L_VAL}
\end{equation}
where $x_{rec}$ represents the reconstructed image, and $x_{target}$ represents the target image. The objective of $VAL$ is to interfere with the vision space by bringing the model's reconstructed image closer to the target image, thereby distancing it from the training samples. We also considered perturbing the latent code extracted by the encoder concerning vision space interference. However, experiments revealed that this method adversely affected model disruption, potentially related to the forward noise addition process. We will provide a detailed explanation in Section \ref{sec:ablation}. Without loss of generalization, We choose the black image as the target image.

\noindent \textbf{Prompt-Robust Augmentation (PRA)}
During the training of text-to-image generation models, proper image-text pairs are typically required and useful for generating high-quality image outputs. However, due to neural networks' strong fitting ability, the brewed adversarial noise may also overfit to perturb the fixed image-text pairs since the prompt information remains unchanged. Attackers can choose various prompts, which we cannot anticipate beforehand. Therefore, adopting fixed text prompts for crafting adversarial perturbations may result in unstable performance and reduced defense effectiveness when encountering different attack prompts. We provide a detailed explanation of the above issues discovered during our experiments in Section \ref{sec:comparison}.
In fact, accurate prompt information is unnecessary for generating adversarial perturbations. The perturbed images should not be related to any specific prompt, 
and should be able to protect against the leakage of sensitive information in different prompts.
To enhance defense stability against various attack prompts, we propose solutions in two aspects. 
\begin{itemize}
    \item At the text input level, to mitigate the overfitting of perturbations to fixed prompt sentences at the text input level, the prompts used for creating adversarial perturbations should ideally be subsets of the majority of common prompts. Consequently, we consider two special characters: 1) the underscore $"_"$, replaced by spaces in CLIP before processing, making it a subset of most prompts, and 2) an empty string $""$, a subset of any string. In Section \ref{sec:ablation}, we showcase the defensive benefits of using these special strings compared to a fixed text prompt. Both special strings ultimately increase image quality degradation and provide interference comparable to the fixed prompt's interference for generated faces.
    \item At the text feature level, we hypothesize that brewing adversarial perturbations with diverse text involved may improve stability against various prompts by back-propagating gradients containing a wider range of text types. Therefore, we propose data augmentation operations on text features. Specifically, we add uniformly distributed random noise to the text features. In Section \ref{sec:ablation}, we also demonstrate the effective improvement brought by adding text feature augmentation to the defense effect.
\end{itemize}
Thus, the text embedding fed into the model is as follows:
% \vspace{-0.05in}
\begin{equation}
    E_{text} = CLIP\left( {prompt} \right), \\
    \tau = E_{text} + N \sim U(-1,1),
    \label{eq:PromptAug}
\end{equation}
where $prompt$ means the input text prompt, $CLIP$ is the model used to convert a text string to embedding, ${N \sim U(-1,1)}$ is the noise in uniform distribution, and $\tau$ is the final text embedding fed into the denoising process.
The objective of $PRA$ is to eliminate the concealed reliance of adversarial perturbations on particular prompts by breaking the connection between the text embedding and feature space. This method enhances the robustness of the generated noise, rendering it effective across a broad spectrum of prompts.

\noindent \textbf{Attention Decoupling Loss}
Before introducing our method, let us briefly review the principle of how the text-to-image diffusion model performs image-text fusion and accomplishes image-text mapping. During the forward process, the model transforms an image into an image latent code approximating a Gaussian distribution through multiple time-step noise additions. Subsequently, the image latent code and text embedding are concatenated into the denoising process, where the image-text fusion takes place. Cross attention mainly undertakes the fusion task, calculating the attention matrix for image and text features already mapped to the unit space. This attention matrix contains crucial image-text correlation coefficients, which are utilized to adjust the text features, enabling the network to further couple the text and image information, as illustrated in Eq. \ref{eq:attention}. The model gradually establishes image-text mapping relationships through iterative training and leverages relevant prompts for personalized creation, highlighting the importance of the image-text fusion module within the diffusion model framework.
% \vspace{-0.05in}
\begin{equation}
    \begin{aligned}
        &M = \operatorname{softmax}\left(\left(W_Q z_{img} \right)\left(W_K z_{text} \right)^T\right), \\
        &\hat{z}_{text} =  M \cdot W_V z_{text}, 
    \end{aligned}
    \label{eq:attention}
\end{equation}
where $M$ is the attention matrix generated by $z_{img}$ and $z_{text}$. $z_{img}$ and $z_{text}$ are the image and text features in unit space. $W_Q, W_K, W_V$ are weights in the cross-attention module.
We focus on the attention matrix within the cross-attention mechanism to interfere with the image-text fusion process and prevent the network from establishing effective image-text mapping relationships. We propose $Attention Decoupling loss (ADL)$ and formulate it as follows:

\begin{equation}
    \begin{aligned}
        % \mathcal{L}_{adl} = \left\| M - \mathbf{I}_n  \right\|_2^2
        \mathcal{L}_{ADL} = \left\| M - M_{target}  \right\|_2^2,     
    \end{aligned}
    \label{eq:adl}
\end{equation}
where $M_{target}$ is a target matrix to disturb ${M}$, which will be endetailed in Sec \ref{sec:ablation}.
$ADL$ aims to draw the attention matrix close to a specific matrix, thereby disrupting the attention matrix's ability to express correct image-text correlation coefficients. By minimizing this loss function, 
the attention information of different channels in $M$ is effectively suppressed, 
causing $z_{img}$ to receive little attention guidance. 
$z_t$ and $z_{img}$ are more likely to maintain high consistency, which leads the model to establish erroneous image-text mapping relationships or even failure in mapping. In summary, the decoupling loss achieves efficient protection by disrupting the fusion process of image and text.

\noindent \textbf{Overall}
In addition to the losses introduced above, which are designed from the vision space and the image-text fusion common unit space, we also considered the adversarial loss corresponding to the conventional DreamBooth training $L_{cond}$, which is introduced in Sec. \ref{sec:preliminaries}. Overall, our combined loss formulation is as follows:
\begin{equation}
    L_{ADAF} = \lambda_{1}L_{VAL} + \lambda_{2}L_{ADL} + \lambda_{3}L_{cond}, 
    \label{eq:L_ADAF}
\end{equation}
where $\lambda$ denote the positive/negative relationship between the specific loss and $L_{ADAF}$. As Eq. \ref{eq:antidb} shows, we generate adversarial noise with maximize the $L_{ADAF}$, so we set ${\lambda_{1}=-1}$, ${\lambda_{2}=-1}$ and $\lambda_{3}=1$.
\section{Experiments}
\subsection{Experimental Setups}

% Table generated by Excel2LaTeX from sheet '主表格v2'
\begin{table*}[]\tiny
  % \vspace{-0.15in}
  \caption{Comparing the defense performance of the proposed methods, promptA="one picture of ajwerq human", promptB="a dslr portrait of rjq person".}
  % \vspace{-0.1in}
  \resizebox{\textwidth}{!}{
    \centering
    \begin{tabular}{c|c|cccccccc}
    \toprule
    \multirow{5}[6]{*}{Dataset} & \multirow{5}[6]{*}{metric} & \multicolumn{1}{c|}{\multirow{2}[2]{*}{method}} & \multicolumn{4}{p{16.18em}|}{token level} & \multicolumn{2}{c|}{\multirow{2}[2]{*}{prompt level}} & \multirow{5}[6]{*}{mean} \\
          &       & \multicolumn{1}{c|}{} & \multicolumn{4}{p{16.18em}|}{a photo of S* person} & \multicolumn{2}{c|}{} &  \\
\cmidrule{3-9}          &       & \multicolumn{1}{c|}{token / prompt →} & S*=sks & S*=t@t & S*=rjq & S*=ajwerq & promptA & \multicolumn{1}{c|}{promptB} &  \\
\cmidrule{3-9}          &       & \multicolumn{1}{c|}{token cosine similarity} & 1     & 0.66  & 0.61  & 0.58  & 0.58  & \multicolumn{1}{c|}{0.61} &  \\
          &       & \multicolumn{1}{c|}{prompt cosine similarity} & 1     & 0.64  & 0.67  & 0.56  & 0.52  & \multicolumn{1}{c|}{0.50} &  \\
    \midrule
    \multirow{12}[8]{*}{CelebA-HQ} & \multirow{3}[2]{*}{BRISQUE↑} & No Defense & 18.32 & 19.92 & 19.96 & 20.60 & 19.09 & 18.05 & 19.32 \\
          &       & ASPL  & \textbf{39.60} & 33.55 & 31.12 & 32.19 & 32.53 & 24.26 & 32.21 \\
          &       & ADAF(ours) & \underline{39.59} & \textbf{36.50} & \textbf{32.89} & \textbf{33.60} & \textbf{33.43} & \textbf{27.03} & \textbf{33.84} \\
\cmidrule{2-10}          & \multirow{3}[2]{*}{FID↑} & No Defense & 154.45 & 139.32 & 141.07 & 142.52 & 150.33 & 150.65 & 146.39 \\
          &       & ASPL  & 496.54 & 479.93 & 451.37 & \textbf{480.42} & \textbf{488.40} & 480.23 & 479.48 \\
          &       & ADAF(ours) & \textbf{504.25} & \textbf{483.75} & \textbf{473.10} & 471.96 & 485.06 & \textbf{488.80} & \textbf{484.49} \\
\cmidrule{2-10}          & \multirow{3}[2]{*}{FDFR↑} & No Defense & 0.02  & 0.00  & 0.01  & 0.01  & 0.01  & 0.01  & 0.01 \\
          &       & ASPL  & 0.80  & 0.45  & 0.30  & \underline{0.31}  & 0.37  & 0.41  & 0.44 \\
          &       & ADAF(ours) & \textbf{0.86} & \textbf{0.58} & \textbf{0.53} & \textbf{0.31}  & \textbf{0.44} & \textbf{0.57} & \textbf{0.55} \\
\cmidrule{2-10}          & \multirow{3}[2]{*}{ISM↓} & No Defense & 0.45  & 0.46  & 0.46  & 0.46  & 0.46  & 0.46  & 0.46 \\
          &       & ASPL  & \textbf{0.08} & 0.13  & 0.14  & \textbf{0.11} & \underline{0.11}  & 0.11  & \underline{0.11} \\
          &       & ADAF(ours) & 0.10  & \textbf{0.12} & \textbf{0.13} & 0.12  & \textbf{0.11}  & \textbf{0.09} & \textbf{0.11} \\
    \midrule
    \multirow{12}[8]{*}{VGGFace2} & \multirow{3}[2]{*}{BRISQUE↑} & No Defense & 21.66 & 19.44 & 18.64 & 19.38 & 17.76 & 17.26 & 19.02 \\
          &       & ASPL  & 38.07 & 36.09 & 35.76 & \textbf{37.65} & 31.37 & 29.85 & 34.80 \\
          &       & ADAF(ours) & \textbf{38.79} & \textbf{37.20} & \textbf{36.40} & 37.28 & \textbf{34.04} & \textbf{32.86} & \textbf{36.10} \\
\cmidrule{2-10}          & \multirow{3}[2]{*}{FID↑} & No Defense & 211.26 & 206.72 & 204.81 & 203.95 & 215.75 & 204.76 & 207.88 \\
          &       & ASPL  & 510.94 & \textbf{499.64} & \textbf{490.22} & 495.50 & 494.81 & 517.40 & 501.42 \\
          &       & ADAF(ours) & \textbf{526.30} & 494.55 & 487.19 & \textbf{511.00} & \textbf{497.94} & \textbf{542.17} & \textbf{509.86} \\
\cmidrule{2-10}          & \multirow{3}[2]{*}{FDFR↑} & No Defense & 0.01  & 0.01  & 0.01  & 0.01  & 0.00  & 0.00  & 0.01 \\
          &       & ASPL  & 0.83  & 0.81  & \textbf{0.59} & 0.74  & 0.74  & 0.63  & 0.72 \\
          &       & ADAF(ours) & \textbf{0.88} & \textbf{0.82} & 0.54  & \textbf{0.78} & \textbf{0.76} & \textbf{0.70} & \textbf{0.75} \\
\cmidrule{2-10}          & \multirow{3}[2]{*}{ISM↓} & No Defense & 0.39  & 0.39  & 0.39  & 0.38  & 0.37  & 0.39  & 0.38 \\
          &       & ASPL  & 0.09  & \textbf{0.05} & \textbf{0.08} & 0.09  & 0.09  & 0.09  & \underline{0.08} \\
          &       & ADAF(ours) & \textbf{0.05} & 0.07  & 0.11  & \textbf{0.08} & \textbf{0.07} & \textbf{0.08} & \textbf{0.08} \\
    \bottomrule
    \end{tabular}%
    }
  \label{tab:tab1}%
  % \vspace{-0.15in}
\end{table*}%

We first illustrate the experimental setups in this part.

\noindent \textbf{Datasets.} Following \textbf{ASPL} \cite{van2023anti}, We conduct experiments on face generation tasks using \textbf{Celeba-HQ} \cite{karras2017progressive} and \textbf{VGGFace2} \cite{cao2018vggface2}.
\begin{itemize}
    \item \textbf{Celeba-HQ} \cite{karras2017progressive} is a high-quality dataset of 30,000 celebrity face images for computer vision tasks. Derived from the original CelebA dataset, it features diverse identities, attributes, and poses in high resolution (1024x1024).
    \item \textbf{VGGFace2} \cite{cao2018vggface2} is a large-scale dataset of over 3.3 million face images from 9,131 distinct identities, spanning a wide range of age, ethnicity, and pose variations. 
\end{itemize}
For each dataset, we choose 50 identities with at least 15 images of resolution above 500 $\times$ 500. For each subject in these datasets, we randomly select 8 images and divide them into two subsets: $S_{1}$(the reference clean image set) and $S_{2}$(the protecting set). Each subset contains 4 images. We then center crop and resize images to resolution 512 × 512. 
% \emph{See supplementary material for details.}

\noindent\textbf{Implementation details.} For adversarial attacks, we adopt the commonly-used Projected Gradient Descent (PGD) \cite{madry2017towards} attack to perform untargeted attacks. We take 50 steps to optimize adversarial perturbations and set the step size to 0.1. The budget is set to  8/255 by default. It takes 5 and 10 minutes for ADAF and ASPL to complete on an NVIDIA A100-SXM4-40GB. For Dreambooth, we train both the text-encoder and U-Net model with a learning rate of $5 \times 10^{-7}$ and batch size of 2. We finetune 1000 iterations with SD1.4 as a pretrained model. In order to evaluate the defense robustness, we simulate diverse prompts that may be used by adversaries from the token level and prompt level. For the token level, the text prompts are as "a photo of $S^{*}$ person", $S^{*} \in \{sks, t@t, rjq, ajwerq\}$. For prompt level, we design two sentences, "one picture of ajwerq person" and "a dslr portrait of rjq person", which differ from the sentence at the token level.

\noindent\textbf{Evaluation metrics.} We aim to protect face ID and prevent adversaries from recreating protected ID after training DreamBooth with these face images. Following \textbf{ASPL}, four metrics are utilized to evaluate the performance: (1) \textit{Face Detection Failure Rate} (FDFR) \cite{van2023anti}, the proportion of images not detected face with RetinaFace detector \cite{deng2020retinaface}. (2) \textit{Identity Score Matching} (ISM) \cite{van2023anti}, the cosine similarity between generated face images and average face features of all clean images, using ArcFace \cite{deng2019arcface}. (3) \textit{BRISQUE} \cite{mittal2012no}, a classic metric for comprehensive evaluation of image quality. (4) \textit{Fréchet Inception Distance} (FID), a measure for computing the distance between the feature vectors of real and generated images. For ISM, lower values mean better performance, the opposite for the rest of the metrics.
    
For ISM, lower values mean better performance, the opposite for the rest of the metrics.

% Table generated by Excel2LaTeX from sheet 'ablationV2'
\begin{table*}[]\tiny
  % \vspace{-0.15in}
  \caption{Ablation study about ADAF.}
  % \vspace{-0.1in}
  \resizebox{0.9\textwidth}{!}{
    \centering
    \begin{tabular}{ccccccccccccc}
    \toprule
    \multicolumn{1}{c|}{\multirow{3}[6]{*}{}} & \multicolumn{2}{c|}{\multirow{2}[4]{*}{vision}} & \multicolumn{2}{c|}{\multirow{2}[4]{*}{attention matrix}} & \multicolumn{4}{c|}{TextAug}  & \multirow{3}[6]{*}{BRISQUE↑} & \multirow{3}[6]{*}{FID↑} & \multirow{3}[6]{*}{FDFR↑} & \multirow{3}[6]{*}{ISM↓} \\
\cmidrule{6-9}    \multicolumn{1}{c|}{} & \multicolumn{2}{c|}{} & \multicolumn{2}{c|}{} & \multicolumn{3}{c|}{prompt aug} & \multicolumn{1}{c|}{feature aug} &       &       &       &  \\
\cmidrule{2-9}    \multicolumn{1}{c|}{} & enc.  & \multicolumn{1}{c|}{dec.} & randn & \multicolumn{1}{c|}{diag.} & None  & "\_"  & \multicolumn{1}{c|}{sentence} & \multicolumn{1}{c|}{aug} &       &       &       &  \\
    \midrule
    \multirow{3}[2]{*}{vision} & \checkmark     &       & \multicolumn{2}{c}{\_} & \multicolumn{3}{c}{\_} & \_    & \textbf{37.59} & 166.56 & 0.00  & 0.41 \\
          &       & \checkmark     & \multicolumn{2}{c}{\_} & \multicolumn{3}{c}{\_} & \_    & 32.90 & \textbf{475.63} & \textbf{0.51} & \textbf{0.11} \\
          & \checkmark     & \checkmark     & \multicolumn{2}{c}{\_} & \multicolumn{3}{c}{\_} & \_    & 34.10 & 160.80 & 0.00  & 0.40 \\
    \midrule
    \multirow{2}[2]{*}{attention matrix} &       &       & \checkmark     &       & \multicolumn{3}{c}{\_} & \_    & \textbf{33.85} & \textbf{478.45} & 0.47  & 0.12 \\
          & \multicolumn{2}{c}{\_} &       & \checkmark     & \multicolumn{3}{c}{\_} & \_    & 33.42 & 475.21 & \textbf{0.50} & \textbf{0.11} \\
    \midrule
    \multirow{3}[2]{*}{prompt aug} & \multicolumn{2}{c}{\_} & \multicolumn{2}{c}{\_} & \checkmark     &       &       & \_    & \underline{33.25} & \textbf{482.72} & 0.48  & \underline{0.11} \\
          & \multicolumn{2}{c}{\_} & \multicolumn{2}{c}{\_} &       & \checkmark     &       & \_    & \textbf{33.62} & \underline{481.43} & \textbf{0.53} & 0.12 \\
          & \multicolumn{2}{c}{\_} & \multicolumn{2}{c}{\_} &       &       & \checkmark     & \_    & 33.13 & 479.10 & \underline{0.49}  & \textbf{0.10} \\
    \midrule
    \multirow{2}[2]{*}{feature aug} &       & \checkmark     & \checkmark     &       & \checkmark     &       &       & \_    & 33.30 & 484.20 & 0.54  & 0.12 \\
          &       & \checkmark     & \checkmark     &       & \checkmark     &       &       & \checkmark     & \textbf{33.84} & \textbf{484.49} & \textbf{0.55} & \textbf{0.11} \\
    \bottomrule
    \end{tabular}%
    }
  \label{tab:tab2}%
  % \vspace{-0.1in}
\end{table*}%

\subsection{Comparison with State-of-the-Art Methods}
\label{sec:comparison}
We first compare our ADAF with ASPL. Notably, ADAF does not rely on fixed text when making anti-perturbation, but ASPL needs a determinate sentence to build text embedding. In order to evaluate the stability of the protection algorithm under DreamBooth models trained with various text prompts, we start with "a photo of sks person" and design several sentences with gradually decreasing similarity from the start, imitating various text prompts that attackers may use during DreamBooth training. We generate 16 pictures with the same prompt as in training. The average scores over DreamBooth-generated images with each method are reported in Table \ref{tab:tab1}. 

As Table \ref{tab:tab1} shows, we can draw several conclusions below: (1) At the token level, as token similarity decreases, ASPL's defensive performance declines, especially for brisque and FDFR. Conversely, our ADAF significantly mitigates this downward trend across metrics. Section \ref{sec:ablation} discusses the optimal defense for both algorithms when S* = sks. (2) At the prompt level, with further reduced text prompt similarity, ASPL's performance declines, while ADAF surpasses ASPL considerably. (3) Our ADAF outperforms ASPL in most metrics within the six simulated attack scenarios. The reported mean values of the last column highlight ADAF's enhanced defense and stability across various prompts. 

\begin{figure}[t]
  \centering
  \includegraphics[width=0.9\linewidth]{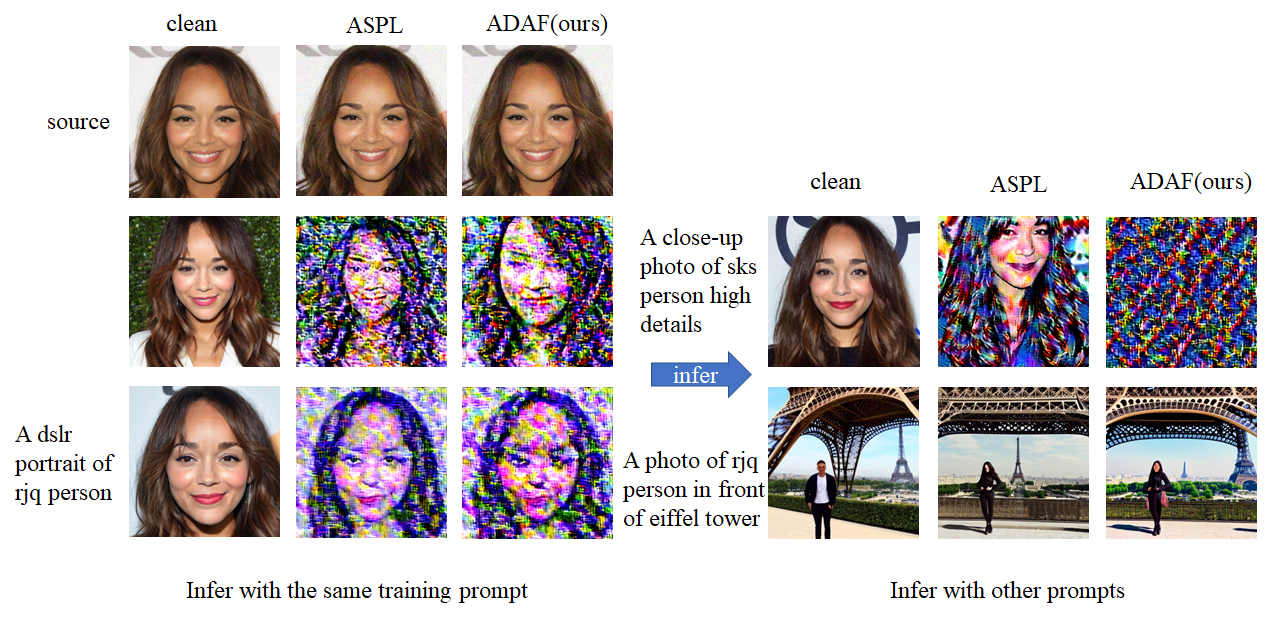}
  % \vspace{-0.1in}
  \caption{Comparison of generated images with the same training prompts and others from disturbed DreamBooth.}
  \label{fig:fig3}
  % \vspace{-0.1in}
\end{figure}

In Figure \ref{fig:fig3}, we assess the defensive performance of ADAF and ASPL. The second row displays the generation results after the attacker trains with the prompt "a photo of sks person", the same as used by ASPL for crafting adversarial perturbations. Both algorithms successfully interfere with the attacker's model, preventing clear image generation of the protected ID, with ADAF showing more pronounced interference than ASPL. The third row shows results after training with the prompt "a dslr portrait of rjq person". The interference level of ASPL-protected images decreases, while ADAF-generated images remain blurry. When attackers use different prompts for diversified creation, as shown in the right of Figure \ref{fig:fig3}, ASPL's defense becomes unstable, whereas ADAF consistently generates blurry or failed faces.

% Table generated by Excel2LaTeX from sheet 'epsV2'
\begin{table}[]\tiny
  \caption{Defense performance with different budgets.}
  % \vspace{-0.1in}
  \resizebox{0.9\linewidth}{!}{
    \centering
    \begin{tabular}{cccccc}
    \toprule
    $\eta$   & FDFR↑ & ISM↓  & BRISQUE↑ & FID↑  & SSIM↑ \\
    \midrule
    0     & 0.01  & 0.46  & 19.32 & 146.39 & 1 \\
    1     & 0.01  & 0.44  & 24.38 & 156.42 & 0.99 \\
    2     & 0.01  & 0.42  & 25.53 & 232.74 & 0.98 \\
    4     & 0.05  & 0.30  & 16.60 & 328.01 & 0.94 \\
    \textbf{8} & 0.55  & 0.11  & 33.84 & 484.49 & 0.83 \\
    16    & 0.85  & 0.07  & 43.32 & 482.32 & 0.63 \\
    \bottomrule
    \end{tabular}%
  }
  \label{tab:tab3}%
  % \vspace{-0.2in}
\end{table}%
\begin{figure}[t]
  \centering
  \includegraphics[width=0.9\linewidth]{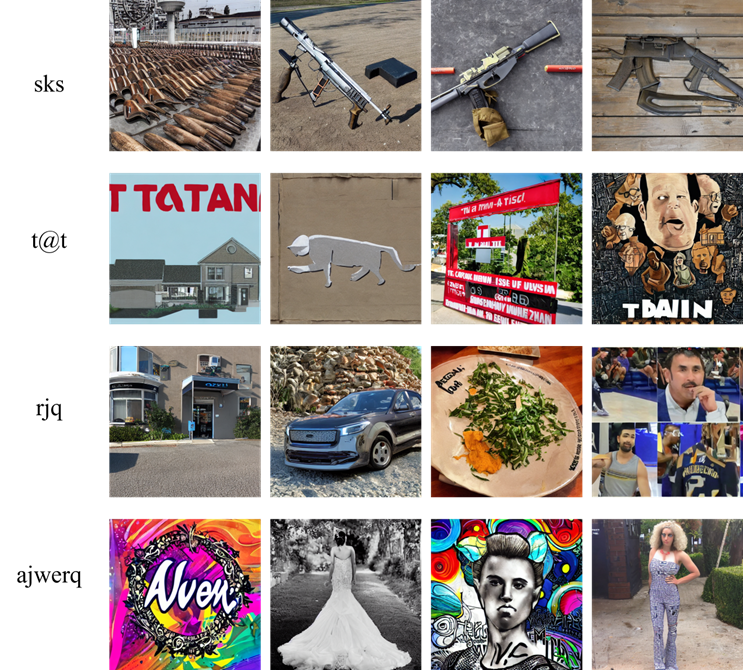}
  % \vspace{-0.1in}
  \caption{Images generated from SD-1.4 with various tokens.}
  \label{fig:fig4}
  % \vspace{-0.1in}
\end{figure}
\begin{figure*}[t]
  \centering
  \includegraphics[width=0.9\linewidth]{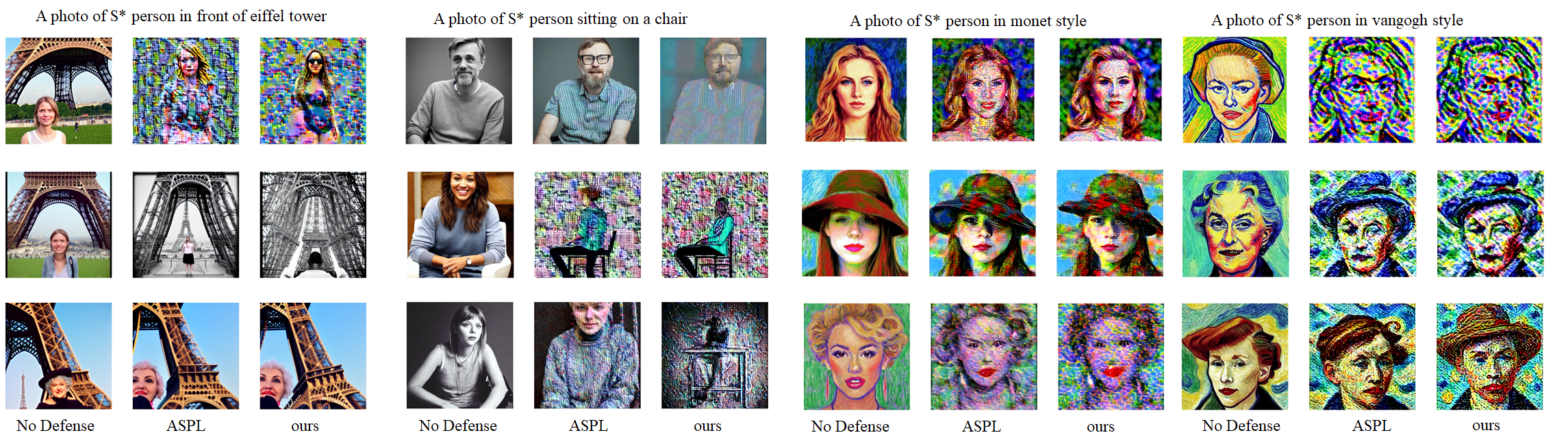}
  % \vspace{-0.1in}
  \caption{More inference results with the training token and various text prompts on CelebA-HQ.}
  \label{fig:fig_celeba}
  % \vspace{-0.05in}
\end{figure*}
\begin{figure*}[t]
  \centering
  \includegraphics[width=0.9\linewidth]{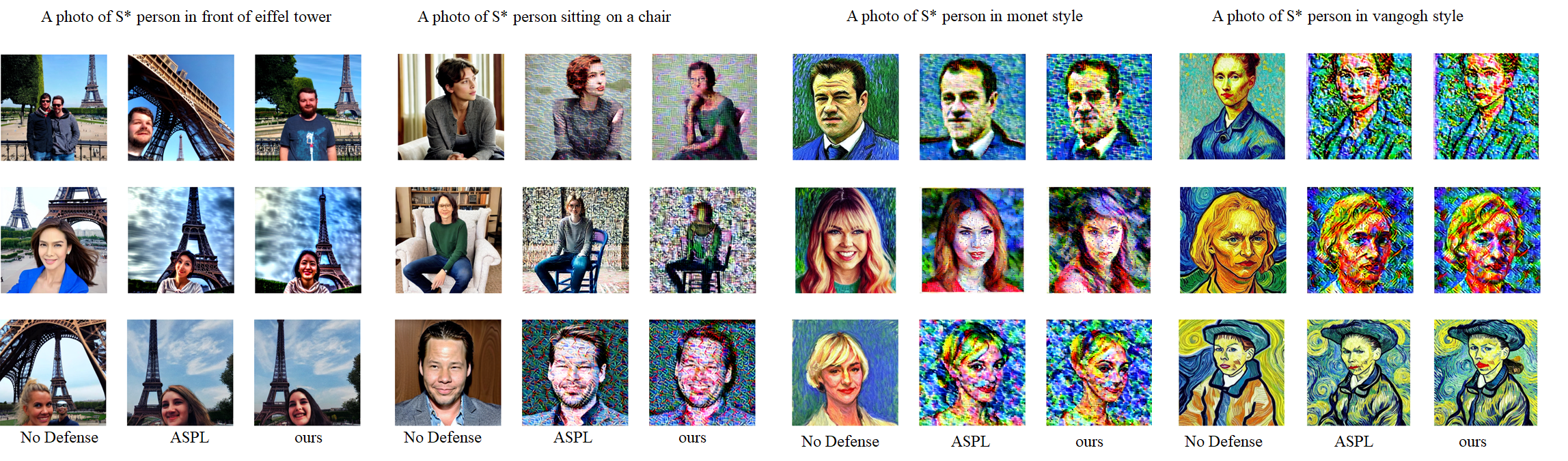}
  % \vspace{-0.1in}
  \caption{More inference results with the training token and various text prompts on VGGFace2.}
  \label{fig:fig_vgg2}
  % \vspace{-0.05in}
\end{figure*}

\subsection{Ablation Study}
\label{sec:ablation}
\noindent\textbf{Vision.} In the vision domain, we examined perturbations applied to both the encoder and decoder outputs. Intriguingly, perturbing the encoder's output solely resulted in a marked decrease in the algorithm's defensive performance. We hypothesize this stems from the forward noise-adding process, transforming any image into Gaussian-distributed noise and deteriorating initial perturbations. Moreover, deteriorated latent codes yield erroneous gradients during denoising, hindering effective adversarial noise generation even with decoder output perturbations. Thus, we focus on designing perturbation loss for the decoder's output in the vision domain.

\noindent\textbf{Cross-attention.} We considered using fixed diagonal matrices and random Gaussian-distributed matrices as perturbations. As shown in Table \ref{tab:tab2}, the two matrix perturbation performances are similar, but diagonal matrices exhibit stronger disturbance for ISM and FDFR, indicating more robust ID protection. Therefore, we employ diagonal matrices for the interference of image-text decoupling.

\noindent\textbf{Text augmentation.} In the text domain, we introduce augmentation at both the input and text feature levels. At the input level, we compared the effects of using a fixed sentence, the common character $"_"$, and the special character "" (null). As the input changes to $"_"$ and "", the image quality-related mean BRISQUE and FID improve. At the same time, the face-related metrics ISM and FDFR remain on par with the fixed sentence approach, indicating the effectiveness of input-level augmentation.
At the feature level, we introduce random noise to interfere with the text features, allowing the gradients collected during perturbation updates to originate from a broader range of text dimensions, further enhancing the defense robustness to various attacker prompts.

\noindent\textbf{Visibility.} We examine the impact of the noise budget ($\eta$) on ADAF's performance. Table \ref{tab:tab3} demonstrates that as $\eta$ increases, image quality metrics, such as BRISQUE and FID, reveal enhanced perturbation performance while the perceptual visibility of the perturbations becomes more apparent (SSIM gradually decreases). Notably, at $\eta$=4, metrics like FDFR and ISM exhibit significant changes, indicating that protective performance is becoming effective. Our defense is already effective at $\eta$=8, with higher noise budgets yielding improved defense performance at the expense of perturbation invisibility.

\noindent\textbf{Special token.} As shown in Table \ref{tab:tab1}, both ASPL and our ADAF exhibit significantly better defense performance when S* = sks. We generated images using the pre-trained model SD-1.4 with the four tokens used in the experiments, as illustrated in Fig. \ref{fig:fig4}. The content for sks primarily features machine gun-like patterns, while other tokens lack correlation. We speculate that sks has already acquired a specific referential meaning in the pretrained model. For unprotected images, the finetuning process does not significantly damage the model's generative capabilities when attackers train with these tokens. However, for protected images, the tokens' inherent referential meaning appears to aid perturbation, considerably weakening the model's generative capabilities after training with these tokens.

\noindent\textbf{Inference with diverse prompts} As depicted in Fig. \ref{fig:fig_celeba} and Fig. \ref{fig:fig_vgg2}, we present the effects of generating images with different prompts from those used in training for attackers. It is evident that, on CelebA-HQ or VGGFace2, attackers can easily repurpose unprotected images. However, the generated image quality and ID information are noticeably degraded after protection through the defense algorithm. Compared to ASPL, our ADAF is more effective in damaging the generated image quality or altering the ID information, indicating the stability of our ADAF against various prompts.
\section{Conclusion}
This paper introduced the Adversarial Decoupling Augmentation Framework (ADAF) to enhance facial privacy protection in text-to-image models by targeting the image-text fusion module. Furthermore, we proposed multi-level text-related augmentations to improve the defense stability of the protection algorithm against various attacker prompts. ADAF comprises three key components: Vision-Adversarial Loss, Prompt-Robust Augmentation, and Attention-Decoupling Loss. Extensive experiments demonstrated ADAF's effectiveness in handling various attacker prompts and outperforming existing methods. ADAF addresses privacy concerns arising from text-to-image models and offers potential future research directions in generative AI privacy protection.

%%%%%%%%% REFERENCES
\bibliographystyle{ACM-Reference-Format}
\bibliography{sample-base}

\clearpage
\end{document}